\documentclass[letterpaper, 10 pt, conference]{ieeeconf}  

\IEEEoverridecommandlockouts                              

\usepackage{graphics} 
\usepackage{epsfig} 
\usepackage{mathptmx} 
\usepackage{times} 
\usepackage{amsmath} 
\usepackage{amssymb}  
\usepackage{enumerate}
\usepackage{multirow}
\usepackage{multicol}
\usepackage{makecell}
\usepackage{booktabs}
\usepackage{cite}
\usepackage{hyperref}
\usepackage[justification=centering]{caption}
\usepackage{marvosym}

\title{\LARGE \bf Feature Map Convergence Evaluation for Functional Module}

\author{
	Ludan Zhang$^{1,2}$, 
        Chaoyi Chen$^{1}$, 
	Lei  He$^{1}${\textsuperscript{\Letter}}, 
	Keqiang Li$^{1}$  \thanks{This work was supported by National Natural Science Foundation of China, Science Fund for Creative Research Groups (Grant No. 52221005), Research and Development of Autonomous Driving Domain Controller and Its Algorithm under Grant 2023Z070, Tsinghua-Toyota Joint Research Institute Inter-disciplinary Program.} \thanks{$^1$ Authors are with Tsinghua University.} \thanks{$^2$ Author is with Nankai University. This work is done during an internship at Tsinghua University.} \thanks{\textrm{\Letter} Corresponding author. E-mail: \href{mailto:helei2023@tsinghua.edu.cn}{helei2023@tsinghua.edu.cn}.}\\
}

\begin{document}
\maketitle
\begin{abstract}
Autonomous driving perception models are typically composed of multiple functional modules that interact through complex relationships to accomplish environment understanding. However, perception models are predominantly optimized as a black box through end-to-end training, lacking independent evaluation of functional modules, which poses difficulties for interpretability and optimization. Pioneering in the issue, we propose an evaluation method based on feature map analysis to gauge the convergence of model, thereby assessing functional modules' training maturity. We construct a quantitative metric named as the Feature Map Convergence Score (FMCS) and develop Feature Map Convergence Evaluation Network (FMCE-Net) to measure and predict the convergence degree of models respectively. 
FMCE-Net achieves remarkable predictive accuracy for FMCS across multiple image classification experiments, validating the efficacy and robustness of the introduced approach. To the best of our knowledge, this is the first independent evaluation method for functional modules, offering a new paradigm for the training assessment towards perception models.
\end{abstract}

\section{INTRODUCTION}

Deep Neural Networks (DNNs) are widely applied in the field of autonomous driving perception, significantly enhancing the vehicle's understanding and response capabilities to its surrounding environment.
The internal architecture of perceptual DNN models is composed of multiple functional modules, such as feature extraction, feature perspective transformation, or feature fusion, and they work in concert to achieve comprehensive environmental understanding.
Currently, the optimization and evaluation of DNN models are largely executed in an end-to-end fashion, treating the entire model as a singular black box. 

The absence of an independent assessment mechanism for intermediate modules further impedes the ability to conduct targeted optimization, which hampers the enhancement of the overall training efficiency of the algorithm. Furthermore, the inherent black-box characteristic of DNNs poses a challenge to interpretability. The opacity surrounding the specific information learned by the neural networks and the decision-making processes they employ makes it challenging to instill trust in these models. This is particularly problematic in the field of autonomous driving, where the need for prudent decision-making is paramount\cite{kuznietsov2024explainable, zablocki2022explainability}.

Recent studies have addressed challenges in training intermediate neural network modules by incorporating auxiliary information. BEVFormer v2 \cite{yang2023bevformer} introduces camera-derived supervisory signals to counteract the interference from the attention module for backbone. Meanwhile, CLIP-BEVFormer \cite{pan2024clip} integrates authentic Bird's Eye View modules with real query interaction, enhancing BEV feature generation and model learning.
Modular networks are based on the principle of decomposing complex neural networks into a series of independent modules, each specialized for processing distinct tasks or features.
viates the problem of catastrophic forgetting, thereby boosting the network's flexibility and scalability.
Inspired by this concept, several practical systems have emerged, including Neural Modular Networks \cite{andreas2016neural}, Capsule Neural Networks \cite{patrick2022capsule}, and PathNet \cite{fernando2017pathnet}, which have achieved groundbreaking advancements in various domains such as image classification, speech recognition, autonomous driving, and facial recognition.
\begin{figure*}
  \centering
  \includegraphics[width=0.9\textwidth]{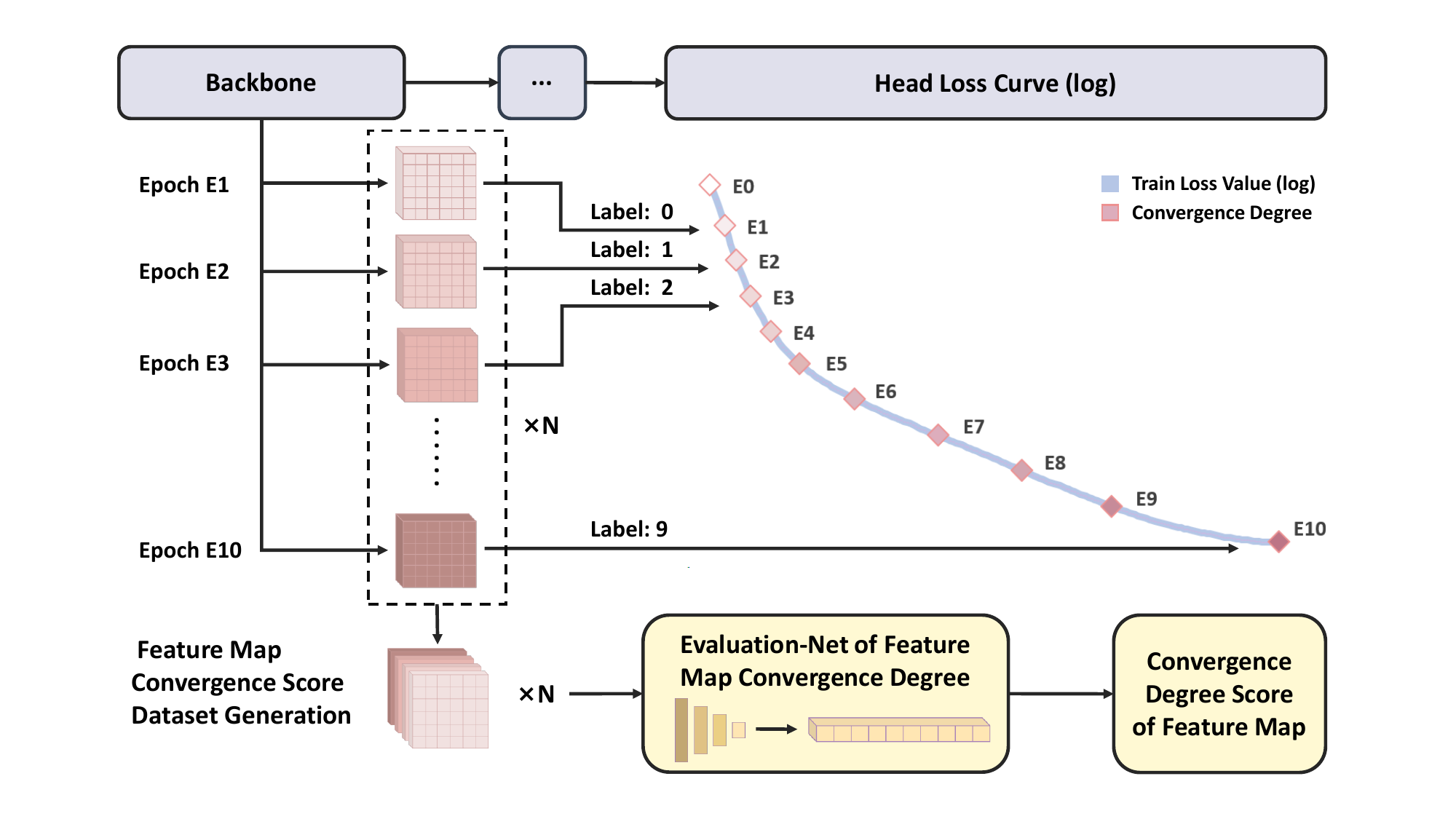}
  \captionsetup{font={small, stretch=1.05},justification=raggedright}
  \caption{\textbf{Feature Map Convergence Evaluation Framework.} This framework is designed to generate a standardized Feature Map Convergence Score (FMCS) for the output feature maps of functional modules, such as the backbone, based on the analysis of the Loss Sequence from the task head. The Feature Map Evaluation Network (FMCE-Net) is trained on the Feature Map Convergence Score Dataset, serving as a tool to assess the maturity of the training of functional modules.}
  \label{fig:method}
  \vspace{-0.6cm}
\end{figure*}

To address the challenge of independently evaluating intermediate modules within end-to-end Deep Neural Network (DNN) models, we adopt the philosophy of modularization to decompose complex DNN models into a series of functional modules based on their specific tasks or purposes. 
The assessment of functional modules is primarily focused on their training maturity, which pertains to the module's effectiveness in extracting output information from upstream modules and providing high-quality input signals to downstream modules. 
Our method for evaluating the training maturity of the module is by assessing the quality of the generated 
 feature maps since the connection between functional modules is forged by the feature maps. Loss is a pivotal metric reflecting a model's prediction accuracy, meanwhile indicative of its training maturity. In image classification, the close connection between the feature extraction module and the classification head renders the assessment of training maturity highly reliable.

Therefore, our objective is to establish a method for evaluating the convergence of output feature maps from the backbone in image classification tasks, as shown in Fig.\ref{fig:method}. Specifically, we first introduce a Convergence Quantification Indicator (CQI), which measures the subtle variations of the Loss Curve. The model is deemed to have converged when the CQI falls below a predetermined threshold $\mu_{CQI}$. Subsequently, we propose a quantification framework to assess the convergence degree of feature maps delivered from the backbone, named as Feature Map Convergence Score (FMCS). Finally, we generate feature maps from $K$ convergence phases on the entire training dataset to construct a feature maps dataset for training the Feature Map Convergence Evaluation Network (FMCE-Net). Extensive experiments are conducted to validate the high accuracy of our evaluation method and visualizations are provided to demonstrate its effectiveness. 

Our contributions can be summarized as follows:
\begin{itemize}
    \item We propose the first independent evaluation method for the training maturity of functional modules within DNNs, in autonomous driving.
    \item We introduce a quantitative criteria for assessing the convergence of models, which traditionally relies on empirical judgment, realized through CQI and FMCS. 
    \item Extensive experiments are developed on image classification tasks, which substantiate the effectiveness and robustness of our method.
\end{itemize}

\section{RELATED WORKS}

\subsection{Statistical Evaluation Method for Feature Maps}
Analyzing feature maps iteratively is a common strategy for comprehensive information extraction and differentiation among feature maps. A Convolution Neural Network (CNN) cache structure was proposed for image classification by calculating cosine similarity between cached feature maps and new inputs \cite{park2018accelerating}. Further insights into CNNs were provided by statistical analyses of feature map disparities across layers and in relation to original images \cite{azam2023using}, shedding light on the inner workings of CNNs. The Area Under the Curve-Receiver Operating Characteristic (AUC-ROC) value was utilized to measure feature map quality, with a random forest regressor for estimation, enhancing the integration of feature maps in human gaze prediction \cite{rahman2017feature}. A method based on information entropy was introduced for quantitatively assessing feature extraction\cite{chen2021feature}. Theoretically robust algebraic topological tools were meticulously applied for an enhanced quantitative evaluation of individual units within CNNs  \cite{zhao2021quantitative}.

\subsection{Visualization Method for Feature Maps}
Feature map heatmaps are widely used for analyzing tasks in perception and image analysis, providing insights into the model representation by highlighting key regions within feature maps \cite{selvaraju2017grad, simonyan2013deep, chattopadhay2018grad}. An Explainable AI (XAI) framework was established to evaluate intermediate layer outputs in lane detection using Gradient-weighted Class Activation Mapping (Grad-CAM) and saliency maps \cite{mankodiya2022od}. The research also assessed the impact of auxiliary tasks on network performance through visualization methods like saliency maps.
Shapley Value-based Class Activation Mapping (Shap-CAM), a novel visualization method, calculates pixel importance using class activation mapping and Shapley values \cite{zheng2022shap}. The self-supervised explainable network (SSINet), was introduced to interpret agent behavior by identifying attention patterns and significant features, as well as analyzing failure scenarios, to produce personalized attention masks for feature map pixels \cite{shi2020self}. Shapley Explanation Networks were designed as a modular component to provide pixel-level Shapley representations for DNN feature maps \cite{wang2021shapley}.

\subsection{Auxiliary Losses in End-to-End Models}
Some researches incorporate auxiliary loss functions within intermediate modules of end-to-end models, providing assistance for independent evaluation and optimization. The concept of end-to-end autonomous driving, leveraging deep neural networks with multiple auxiliary tasks, was explored in a study from \cite{wang2019end}. In the realm of monocular 3D object detection, a method was proposed in \cite{kim2022boosting} that enhances detection capabilities by training the network with object-centric auxiliary depth supervision. A novel approach was presented \cite{zheng2023gconet+} with the introduction of GCoNet+, a model designed for co-salient object detection. This model incorporates a recurrent auxiliary classification module, a confidence enhancement module, and employs group-based symmetric triplet loss, all of which contribute to heightened accuracy in detection tasks. 

\section{PROBLEM FORMULATION}
Our core task is to define a quantitative metric for evaluating the convergence of feature maps, known as the \textbf{Feature Map Convergence Score (FMCS)}. Subsequently, we design a neural network called \textbf{Feature Map Convergence Evaluation Net (FMCE-Net)} to predict the FMCS.

Let \( X \) be an image dataset comprising a training subset \( X_{\text{train}} \) and a test subset \( X_{\text{test}} \). Each image \( x_n \in X \) is paired with a corresponding class label \( y_n \). We employ a DNN model \( D \) to classify these images, mapping each input image to its predicted class label. The model's prediction for an image \( x_n \) is expressed as \( \hat{y}_n = D(x_n) \), where \( \hat{y}_n \) represents the estimated class label.

The model $D$ can be considered as a multi-layer composite function and we only focus on the part that generates the Feature Map before the average pooling and linear layer, so $D$ can be divided into 
\[D(x_n) = F(C(x_n)),\] 
where $F$ represents Global Average Pooling (GAP) and Fully Connected (FC)  layers, and $C(x_n)$ is the backbone that generates the feature map.

Without considering over-fitting cases, as the number of training epochs increases, the convergence of the model improves.
By analyzing the loss curve, the training process is divided into $K$ convergence phases, labeled from 1 to $K$. We identify K epoch markers $E_1, \ldots, E_K$ within each convergence phase and extract the corresponding feature maps $ C_{E_k} $ generated at epoch marker. The Feature Map Convergence Score (FMCS) is then associated with the convergence phase label, and the FMCS of  $C_{E_k}$  is assigned the value $k$, as
\[\text{FMCS} (C_{E_{k}}(x_{n}))  = k.\]

Given feature maps as input, we train the FMCE-Net, denoted as \( E \), to achieve an accurate prediction of FMCS. The process can be represented as follows:
\[ \text{FMCS}_{predicted} = E(C_{E_{k}}(x_{n})) \]

\section{METHOD}
In this paper, the feature map convergence evaluation method for functional modules is validated on image classification tasks, since the clear modular structure and short training pipeline ensure a credible evaluation of training maturity. We define the \textbf{Convergence Quantification Indicator (CQI)} and the Feature Map Convergence Score (FMCS) to measure the convergence of entire model and feature maps from the backbone, respectively. Finally, we construct a dataset that incorporates these feature maps to train the Feature Map Convergence Evaluation Net (FMCE-Net).

\subsection{Original Task}
Image classification models generally consist of three key components:
(1) Feature Map Extraction (backbone): This involves the use of convolutional layers to extract local features from images. (2) Feature Map Integration(GAP + FC layers): The information from the feature maps is consolidated, which also helps in reducing the feature representation's dimensionality. (3) Classification Head (FC + Softmax layers): The high-level features are then mapped onto the classification labels by FC layers. Softmax function calculates the output probabilities for each class.

We first conduct an image classification on dataset $X$, termed the \textit{Original Task}, to distinguish it from subsequent feature map evaluation model. The maximum number of training epochs is set to \( M \), for each training epoch \(m\), we record the classification Loss as $L_{m}$ and save the corresponding weights as $W_{m}$, comprising the Loss Sequence $\{L_{m}\}$ and the Weight Sequence $\{W_{m}\}$ respectively. The Loss Curve graphically depicts the progression of the Loss Sequence over the course of a model's training, as shown in Fig \ref{fig:4curve}.
\begin{figure}[ht]
  \vspace{-0.3cm}
  \centering
  \includegraphics[width=0.45\textwidth]{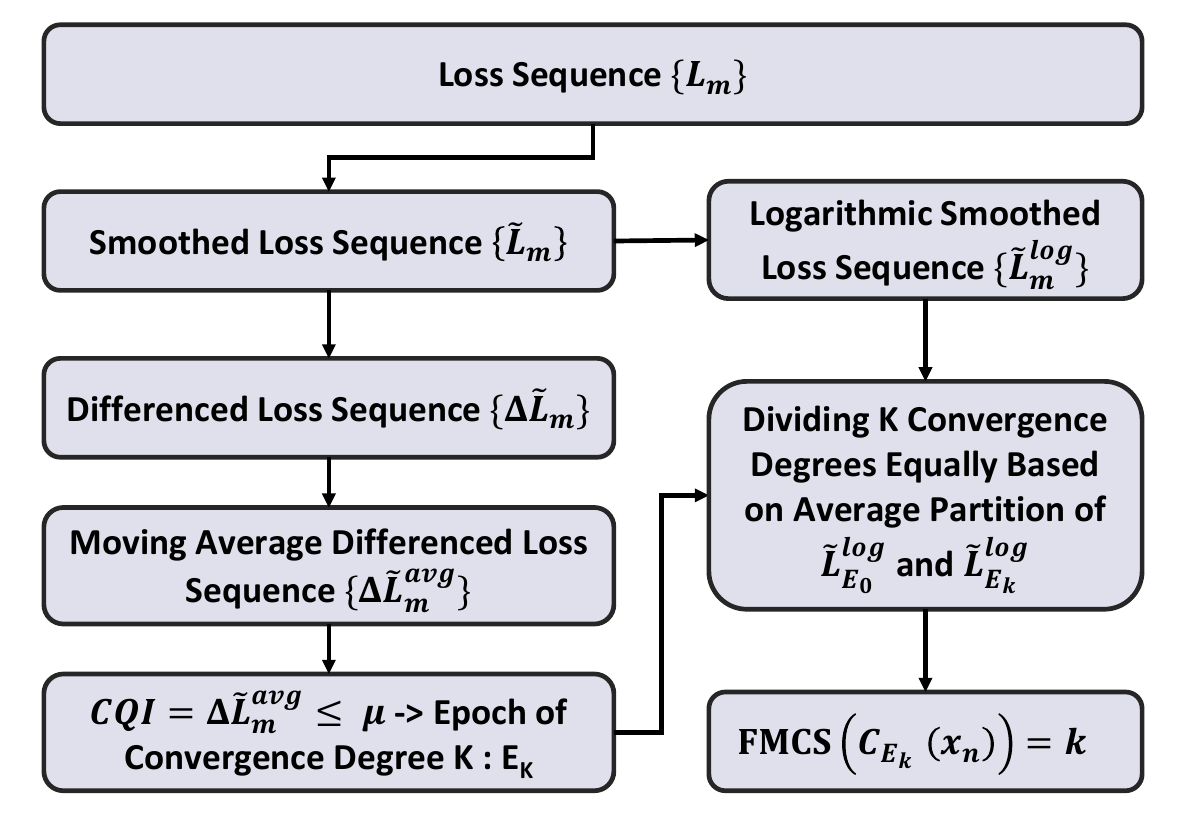}
  \captionsetup{font={small,stretch=1.1},justification=raggedright}
  \caption{\textbf{CQI and FMCS Generation Flowchart}. The CQI serves as a quantitative measure to assess the fluctuation extent of the Loss Sequence. Throughout the training process, \textit{K} epoch markers are identified that represent the convergence phases that are evenly divided, with FMCS 1, \ldots, $K$ for feature maps corresponding to $E_1$, \ldots, $E_K$. } 
  \label{fig:CQI and FMCS}
  \vspace{-0.6cm}
\end{figure}
\subsection{Convergence Quantification Indicator (\(CQI\)) and Convergence Criteria \(\mu_{CQI}\)} \label{CQI}

\begin{figure*}[ht]
  \centering
  \includegraphics[width=1\textwidth]{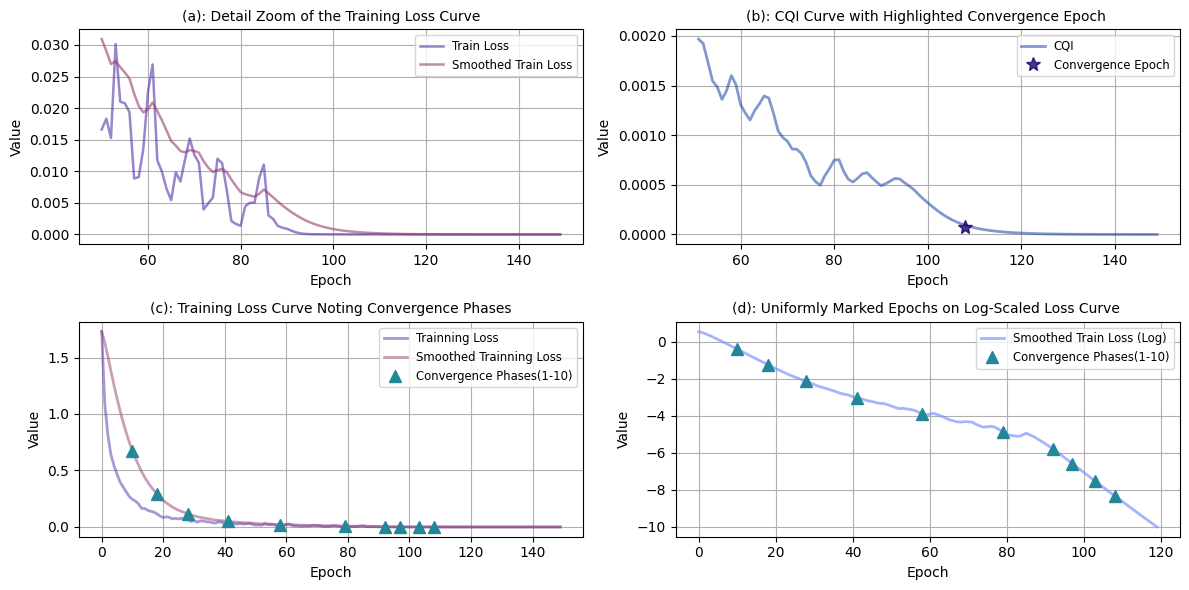}
  \captionsetup{font={stretch=1},justification=raggedright}
  \caption{\textbf{In-Depth Visualization of Training Loss Curves and Epoch Markers.} This figure shows the analysis of Convergence Metrics during ResNet-50 Training on the Mini-ImageNet Dataset over 150 Epochs. (a) presents the original and smoothed Loss Curve observed in a detailed view of the fluctuations around local region of convergence. (b) illustrates the trend of CQI around convergence threshold. 
(c) demonstrates a schematic representation epoch markers to signify convergence phases. 
(d) shows the uniform division of 10 convergence phases on logarithmically smoothed Loss Curve. }
  \label{fig:4curve}
  \vspace{-0.6cm}
\end{figure*}
The Loss Sequence and Loss Curve are indispensable for gauging the convergence of a training model. It is widely accepted  that when the Loss Curve exhibits minimal fluctuations and approaches a horizontal line, it indicates that the model has converged. However, the absence of an exact quantitative measure to confirm this convergence necessitates the creation of a metric that assesses the degree of flatness in the Loss Curve. Upon reviewing Figure \ref{fig:4curve}(c), one might conclude that the model has potentially converged after 80 epochs due to the nearly linear appearance of the Loss Curve. However, a closer look at Figure \ref{fig:4curve}(a) reveals that minor fluctuations are still present. True convergence is only declared when these fluctuations are below a defined threshold, and we will define the Convergence Quantification Indicator (CQI) and\textbf{ CQI convergence threshold\textbf{ \( \mu_{CQI} \)} }as following.

Initially, we apply a smoothing filter to the raw Loss Sequence ${L}=\{L_m\}_{m=1}^{M}$, yielding the smoothed Loss \( \widetilde{L}_m \) for epoch \( m \), to mitigate random fluctuations and better discern the trend. This smoothed sequence is computed using the weighted average:
\[ \widetilde{L}_m = \alpha \cdot {L}_{m-1} + (1 - \alpha) \cdot L_m \]
Here, \( 0 < \alpha \leq 1 \), commonly set to 0.85, is the smoothing parameter. Fig \ref{fig:4curve}(a) (c) provides a comparative illustration of the Loss Curve before and after smoothing. 

Subsequently, we calculate the first-order difference of the smoothed sequence, indicating the change in Loss between successive epochs:
    \[ \Delta \widetilde{L}_m = \widetilde{L}_m - \widetilde{L}_{m-1} \]

To evaluate the stability of these changes, we consider the absolute differences and apply a moving average over a window of \( B \) epochs to derive a measure of variation:
\[ \Delta \widetilde{L}_{m}^{avg} = \frac{1}{b} \sum_{b=1}^{B} |\Delta \widetilde{L}_{m-b+1}| \]
For \( m < B \), \( B \) is replaced by \( m \). This average variation \( \Delta \widetilde{L}_{m}^{avg} \) serves as our Convergence Quantification Indicator (CQI), quantifying the fluctuation extent of the Loss Sequence. Fig.\ref{fig:4curve}(b) delineates the progressive changes in CQI.

Finally, we define a CQI convergence threshold \( \mu_{CQI} \), which determines sufficiently small fluctuation. If the moving average of absolute differences \( \Delta \widetilde{L}_{m}^{avg} \) is less than or equal to \( \mu \), we consider the Loss Sequence to have converged, denoted by:
    \[ CQI_m = \Delta \widetilde{L}_{m}^{avg} \leq \mu_{CQI}  \]

\begin{table*}[h]
\centering
\renewcommand\arraystretch{1}
\captionsetup{font={small,stretch=1.2},justification=raggedright}
\caption{\textbf{Epoch Markers of 10 Convergence Phases.} This table shows training information on different dataset and backbone combinations, including epoches, convergence thresholds, and accuracy level achieved on test sets. Epoch markers of 10 convergence phases are extracted as the basis for generating FMCS-Dataset. }
\scalebox{1}{
\begin{tabular}{ccccccc}
\toprule
\textbf{Dataset} & \textbf{Classes} & \textbf{Model} & \textbf{Epochs} & {\textbf{$\mu_{CQI}$}}&\textbf{Phases of Convergence} &\makecell[c]{\textbf{\textit{Original Task} }\\ \textbf{Top-1 Acc}} \\

\midrule
\multirow{2}{*}{\textbf{MNIST} }
& \multirow{2}{*}{10}& ResNet-50 & 100&1e-4 & {[7, 13, 20, 29, 40, 50, 60, 65, 70, 74]} & 0.9926  \\
& & ShuffleNet V2 &100&1e-4& {[7, 12, 18, 25, 33, 43, 54, 65, 71, 76]} & 0.9927   \\
\midrule
\multirow{2}{*}{\textbf{CIFAR-10} }
&\multirow{2}{*}{10} &ResNet-50 &150&1e-4& {[11, 19, 29, 42, 59, 80, 93, 98, 104, 109]} & 0.9158   \\
&& ShuffleNet V2 &150&1e-4&[11, 20, 30, 45, 65, 83, 95, 101, 107, 113] & 0.8734   \\
\midrule
\multirow{2}{*}{\makecell{\textbf{Mini-}\\\textbf{Imagenet}}} 
& \multirow{2}{*}{100}&ResNet-50 & 150&1e-3&[10, 18, 26, 36, 48, 61, 76, 94, 114, 145]& 0.7860  \\
&& ShuffleNet V2 &250&1e-3&[9, 15, 23, 35, 50, 70, 94, 124, 161, 215]& 0.7133 \\
\bottomrule
\end{tabular}
}
\label{tab:FMCS Dataset}
\vspace{-0.6 cm}
\end{table*}

Through this process, we can quantitatively evaluate whether a model has converged, that is, whether the Loss values have reached a sufficiently stable level, by calculating CQI. The process is diagrammatically represented as a flowchart in the left half of Fig.\ref{fig:CQI and FMCS}.

\subsection{Feature Map Convergence Score (FMCS)}\label{FMCS}
After establishing a quantitative criterion for convergence, we can accurately determine at which epoch the model converges, allowing us to evenly divide the convergence phases. Assuming that the entire training phase is divided into $K$ convergence phases, labeled as 1, ..., $K$, the Kth phase represents the final convergence phase. We define a Feature Map Convergence Score (FMCS) that aligns with the labels of the convergence phases, meaning the FMCS takes on values from 1 to $K$, with $K$ representing the highest degree of convergence. Below, we will describe the method for evenly dividing the convergence stages and the approach to obtaining the corresponding feature maps for $K$ FMCS values.

Firstly, we normalize the scale of smoothed Loss values by applying a logarithmic-linear transformation to the Smoothed Loss Sequence:
\[ \widetilde{L}_{m}^{log} = \log(\widetilde{L}_m) \]
where \( \widetilde{L}_m \) is the smoothed Loss at epoch \( m \). During the training process, the Loss values typically decrease in an exponential manner. Therefore, calculating the logarithm of the Loss Sequence can linearize the exponential function, which aids in our understanding of the convergence process and enables uniform segmentation.

Secondly, To delineate \( K \) convergence phases, it is required to select the Loss values corresponding to \( K \) epochs as the markers for these phases. Initially, the convergence epoch is determined by setting the first epoch where \(CQI\) falls below the convergence threshold \( \mu_{CQI} \) as \( E_K \), as shown in Fig.\ref{fig:4curve}(b). The epoch corresponding to the maximum value in the sequence \( \{L_m\} \) is denoted as \( E_0 \) (typically the first epoch), and it serves as the baseline for calculating the Loss difference with the convergence epoch \( E_k \), given by 
             \[\textit{G}( \widetilde{L}^{log}) = | \widetilde{L}_{E_K}^{log} -\widetilde{L}_{E_0}^{log} |\]

Thirdly, the decrease for each convergence phase \( \Delta G(\widetilde{L}^{log}) \), is determined by dividing the total change by \( K \): \( \Delta G(\widetilde{L}^{log}) = {G(\widetilde{L}^{log})}/{K} \). The set of epochs \( \{E_{k}\}_{k=1}^{K} \), as the \textbf{epoch markers}, demarcate the \( K \) evenly spaced convergence phases. For each epoch $m$ \(( m = 1, 2, \ldots, M) \), we find the epoch \( E_k \) where the cumulative value change  $| \widetilde{L}_{E_k}^{log} -\widetilde{L}_{E_0}^{log} |$ equals approximately \( k \times \Delta G(\widetilde{L}^{log}) \) by summing changes in the Loss Sequence until the condition is satisfied. A visualization example is provided in Fig.\ref{fig:4curve}(c) and (d)

Fourthly, We assign a convergence score \( k \) to the feature maps \( C(x_{n}) \) generated at epoch \( E_k \), denoted as \( \text{FMCS} (C_{E_{k}}(x_{n})) = k \). This score quantifies the degree of convergence for the feature maps at that epoch.

Through the above process, we divide the training process into $K$ convergence phases based on the Loss Sequence, and identify the epoch markers \( \{E_{k}\}_{k=1}^{K} \), that represent these $K$ convergence phases. Additionally, we define FMCS 1 to $K$ for feature maps at epoch markers $E_1$ to $E_K$. The process is diagrammatically represented as a flowchart in the right half of Fig.\ref{fig:CQI and FMCS}.

\subsection{Feature Map Convergence Score Datasets}
\label{FMCS_dataset}
We will generate the \textbf{Feature Map Convergence Score Datasets (FMCS-Dataset)} from the image classification task, also denoted as the \textit{Original Task}. This process involves leveraging the weights \( W_{E_k} \) of these epoch \( \{E_{k}\}_{k=1}^{K} \), for inferencing on the training dataset \( X_{\text{train}} \) of the \textit{Original Task} to obtain feature maps of $K$ convergence phases. For each input sample \( x_n \) in \( X_{\text{train}} \), the corresponding feature maps \( C_{E_k}(x_n) \) are generated and subsequently saved. The Feature Map Convergence Score (FMCS) associated with each feature map is used as the label. Assuming there are \( N \) samples in \( X_{\text{train}} \), this post-inference step results in \( K \times N \) samples. Collectively, these samples constitute the FMCS-Dataset, which is then utilized to train the model for evaluating the convergence degree of feature maps.

\begin{figure*}[!htp]
  \centering
  \includegraphics[width=0.8\textwidth]{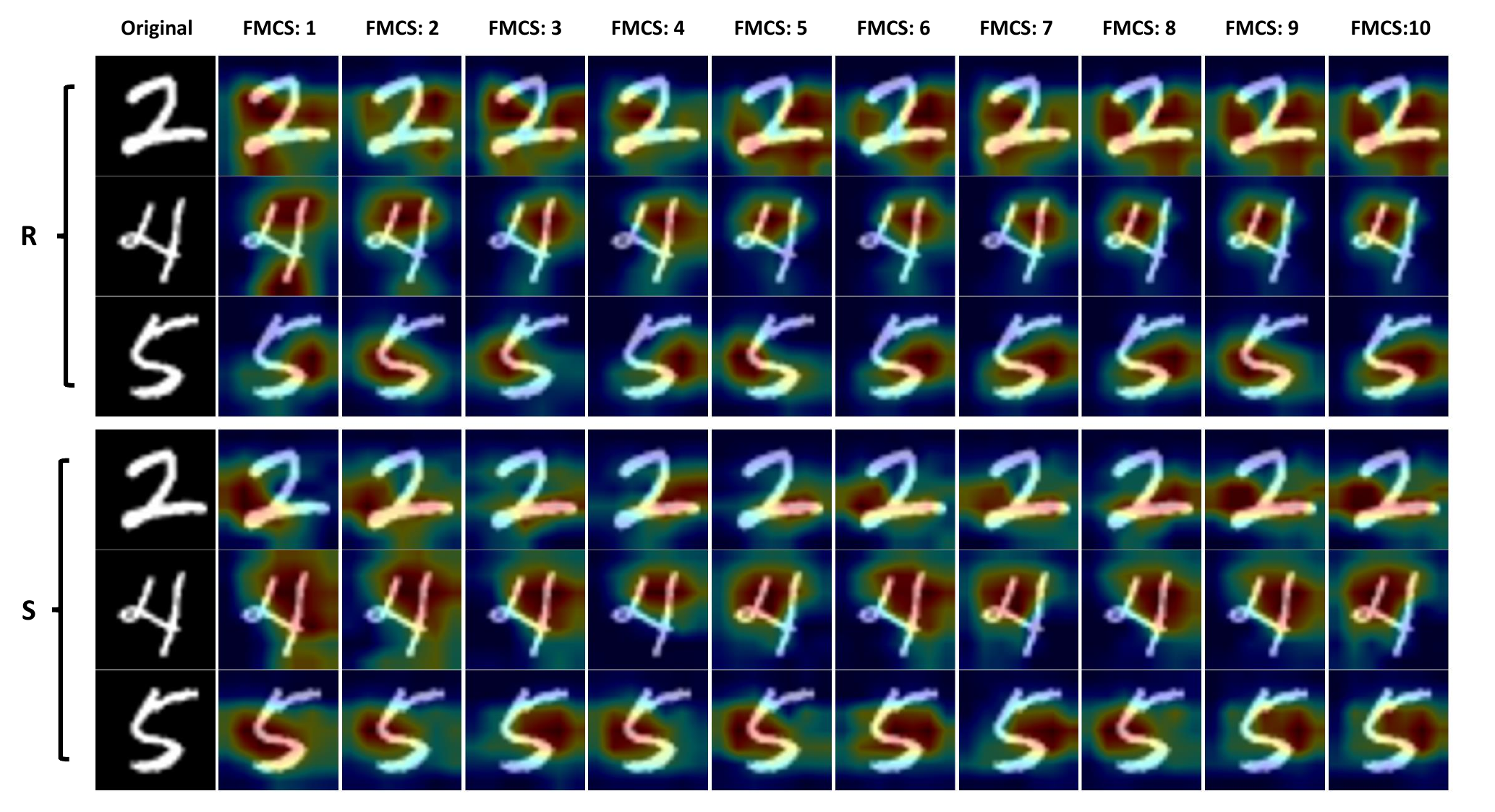}
  \captionsetup{font={small, stretch=1.1},justification=raggedright}
  \caption{\textbf{
  Grad-CAM Heatmaps Across 10 Convergence phases of  MNIST}. Visualizations of feature maps, derived from ResNet-50 (R) and ShuffleNet v2 (S) backbones, same for the two images below. As the FMCS values escalate, the regions of high contribution increasingly localize, indicative of improved feature extraction efficiency. }
  \label{fig:mnist}
  \vspace{-0.3cm}
\end{figure*}

\subsection{Feature Map Convergence Evaluation Network }
\label{EMCE-Net}
Considering our input feature maps annotated with FMCS labels, for effectively
 evaluation of convergence, a natural approach is to leverage Convolutional Neural Networks (CNNs) for predictive classification. CNN is renowned for their capacity to autonomously learn and extract complex hierarchical features from raw data inputs, especially in image tasks. CNNs are inherently modular, allowing them to be readily integrated into existing machine learning pipelines. For embedding into the original model without compromising training efficiency, the network should be designed as a simple architecture.  

To this end, the architecture of Feature Map Convergence Evaluation Network  (FMCE-Net) is tailored to the dimensions of the feature map tensor and typically comprises 2-3 convolutional layers, each succeeded by a MaxPooling layer and a ReLU activation function. Following convolutional layers, the network incorporates FC layers that further process these feature information to culminate in the generation of the FMCS score prediction vector.

\section{EXPERIMENT}

To evaluate the performance of FMCE-Net, we conduct experiments across a range of image classification tasks. Training on \textit{Original Task} allow us to obtain the indicators for judging convergence and the $K$ epoch markers' feature maps, which will then be used to create the FMCS-Dataset. Subsequently, FMCE-Net will be trained on multiple FMCS-Datasets to verify the effectiveness and robustness of the method. Our entire experiment are structured into three processes, followed by an in-depth analysis subsection.

\subsection{Original Task (Image Classfication) Training}
We conduct training on three distinct datasets for image classification tasks: MNIST, CIFAR-10, Mini-Imagenet. And utilizing two commonly employed backbone architectures: ResNet-50 and ShuffleNet v2, as backbone for feature extraction. Throughout the training process, we record the Loss values and saved the weights at each epoch.

\begin{table}
\centering
\captionsetup{font={small,stretch=0.9},justification=raggedright}
\caption{\textbf{FMCS Prediction Result}}
\scalebox{1}{
\begin{tabular}{ccccc}
\toprule
\textbf{FMCS Dataset} & \makecell[c]{ \textbf{Accuracy}} & \makecell[c]{ \textbf{Precise}} &\makecell[c]{\textbf{Recall}}&\makecell[c]{\textbf{F1-Score}}\\
\midrule
\textbf{MNIST-R}&0.9994&0.9994&0.9994&0.9994\\

\textbf{MNIST-S}&0.9942 &0.9943 &0.9942 &0.9942 \\

\textbf{CIFAR-R}&0.9409&0.9418&0.9409&0.9404\\

\textbf{CIFAR-S}&0.9200&0.9276&0.9200&0.9201\\
\textbf{Imagenet-R}&0.9989&0.9989&0.9989&0.9989\\

\textbf{Imagenet-S}&0.9987&0.9987&0.9987&0.9987\\
\midrule
\textbf{Averange}&\textbf{0.9754}&\textbf{0.9768}&\textbf{0.9754}&\textbf{0.9753}\\
\makecell[c]{\textbf{Standard deviation}}&\textbf{0.0354}&\textbf{0.0354}&\textbf{0.0354}&\textbf{0.0353}\\

\bottomrule
\end{tabular}
}
\label{tab:FMCS prediction}
\vspace{-0.6cm}
\end{table}

\begin{figure*}[!htp]
  \centering
  \includegraphics[width=0.8\textwidth]{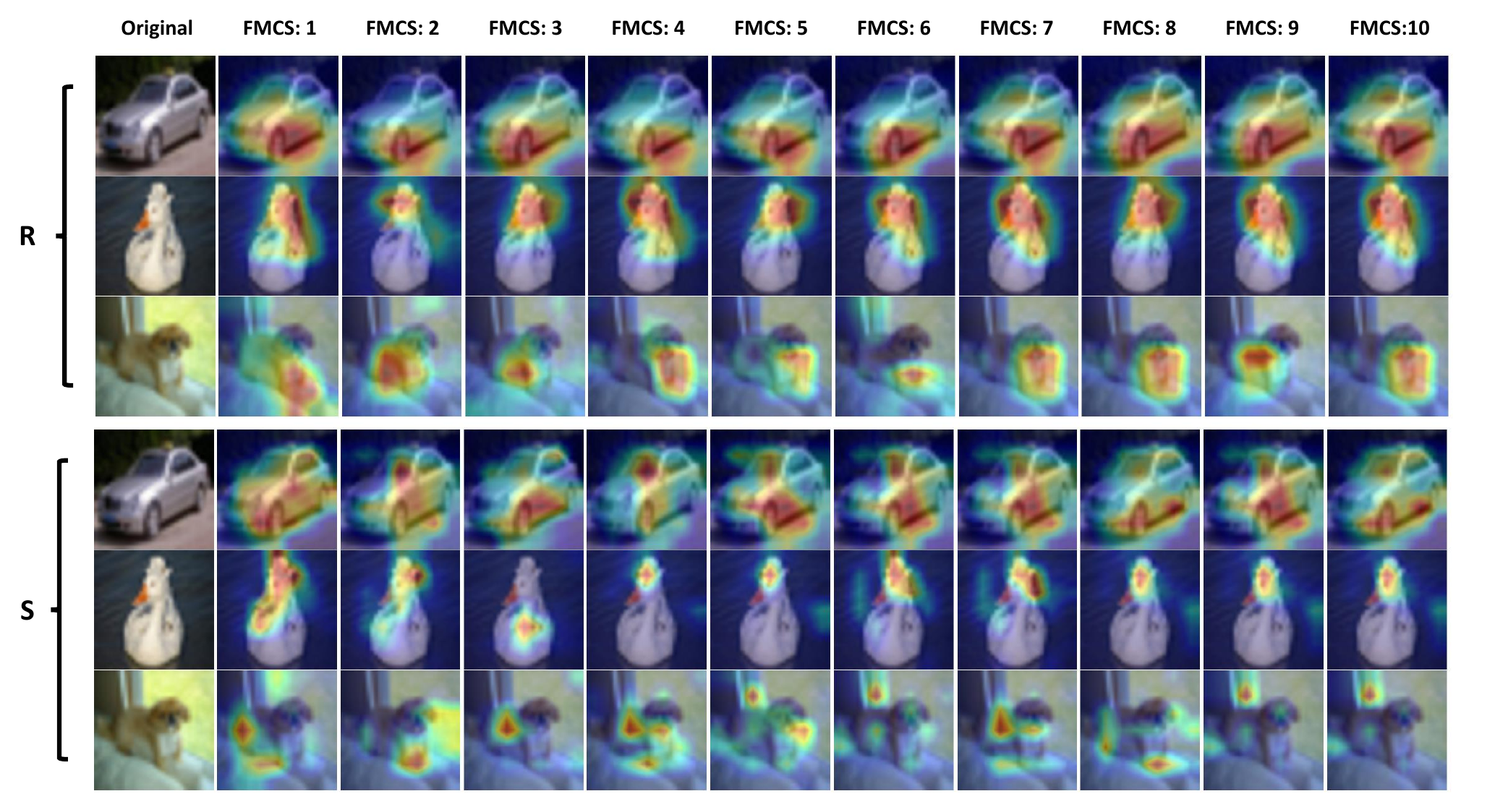}
  \captionsetup{font={small,stretch=1.1},justification=raggedright}
  \caption{\textbf{Grad-CAM Heatmaps Across 10 Convergence phases of CIFAR-10. } Evidently, ResNet-50 does not exhibit a concentrated focus, and ShuffleNet erroneously focuses on the background when extracting features from images labeled with 'dog'. These lead to a relatively lower FMCS predictive accuracy.}. 
  \label{fig:cifar}
  \vspace{-0.9cm}
\end{figure*}

To ensure a consistent experimental setup as much as possible, images of each datasets are resized to a uniform shape of \( 224 \times 224 \) pixels. In particular, since the MNIST dataset consists of grayscale images, we replicated the pixel values across three channels to match the input dimensions of RGB images. Then normalized the image pixel values and applied data augmentation techniques to enhance the robustness of our models. All models employed the cross-entropy function as the loss function to evaluate the discrepancy between the predicted and true categories. For each experiment, we set the batch size to 256 and initialized the learning rate at 0.001. The optimization was performed using the Adam optimizer, which employs a cosine annealing strategy for learning rate decay. The number of training epochs required to reach convergence varied across different dataset-backbone combinations, and the specific details are presented in TABLE \ref{tab:FMCS Dataset}. All training and inference experiments were carried out on a single NVIDIA A100 40G GPU. 

\subsection{Feature Map Convergence Score Dataset Generation}
After obtaining the Loss Sequence $\{L_{m}\}$ and weight $\{W_{m}\}$ for each dataset-backbone combination, we set the Convergence Quantitative Indicator threshold values to $\mu_{CQI} = 1e-3 $  and $\mu_{CQI} = 1e-4 $ for datasets with 100 classes and 10 classes, respectively. With the number of convergence phases set to $K = 10$, we identified the 10 epoch markers $\{E_k\}_{k=1}^{10}$ for FMCS 1-10, in the steps detailed in METHOD \ref{CQI} and \ref{FMCS}. We then loaded the weights from these ten epochs $\{W_{E_k}\}_{k=1}^{10}$ for concentrated inference on the training set, yielding the feature maps corresponding to each of the 10 convergence phases. This process culminated in the creation of a dataset designed to evaluate the convergence degree of feature maps. The resulting feature maps had dimensions of $2048 \times 7 \times 7$ for ResNet-50 and $1048 \times 7 \times 7$ for ShuffleNet v2. For the MNIST, CIFAR-10, and Mini-Imagenet datasets, the number of images in the training sets are denoted as $N_M = 60000$, $N_C = 50000$, and $N_I = 48000$, respectively. When generating feature maps corresponding to 10 distinct convergence phases, the resulting datasets consist of $10 \times N_i$ feature maps for each dataset, where $i$ corresponds to the dataset indicator (M for MNIST, C for CIFAR-10, I for Mini-Imagenet).
TABLE \ref{tab:FMCS Dataset} presents the specific details of generating the FMCS Dataset by applying two different combinations of backbones on three datasets.

\subsection{Feature Map Convergence Evaluation Networks}
The FMCE-Net is CNN architecture and the layer depth correlates with the dimensions of the feature maps produced by the backbone. For instance, when utilizing the ShuffleNet backbone, which generates feature maps of size $1024 \times 7 \times 7$, the FMCE-Net consists of two successive 3×3 convolutional layers, each are followed by MaxPooling and ReLU layers, in order to halving the spatial dimensions and the output channels. This results in a tensor of size 
$256 \times 1 \times 1$, that is then passed to FC layer for feature integration. These processes conclude with the generation of a 10-dimensional vector representing the predictive probabilities for ten different Feature Map Convergence Scores (FMCS). In contrast, when employing the ResNet50 backbone that generates feature maps of size $2048 \times 7 \times 7$, an additional convolutional layer is integrated into the FMCE-Net, which serves to reduce the channel dimensions equal with ShuffleNet's input. Each FMCS-Dataset is meticulously divided into training and testing subsets, maintaining a ratio of 3:1, Furthermore, the distribution ensures that each label has an identical number of feature maps both in the training and testing set.

The predictive accuracy of FMCE-Net for FMCS is evaluated using metrics borrowed from image classification tasks. Specifically, the assessment is conducted through four key metrics: accuracy, precision, recall, and the F1-score. These metrics provide a comprehensive evaluation of the performance capabilities of FMCE-Net.

\begin{figure*}[htp]
  \centering
  \includegraphics[width=0.8\textwidth]{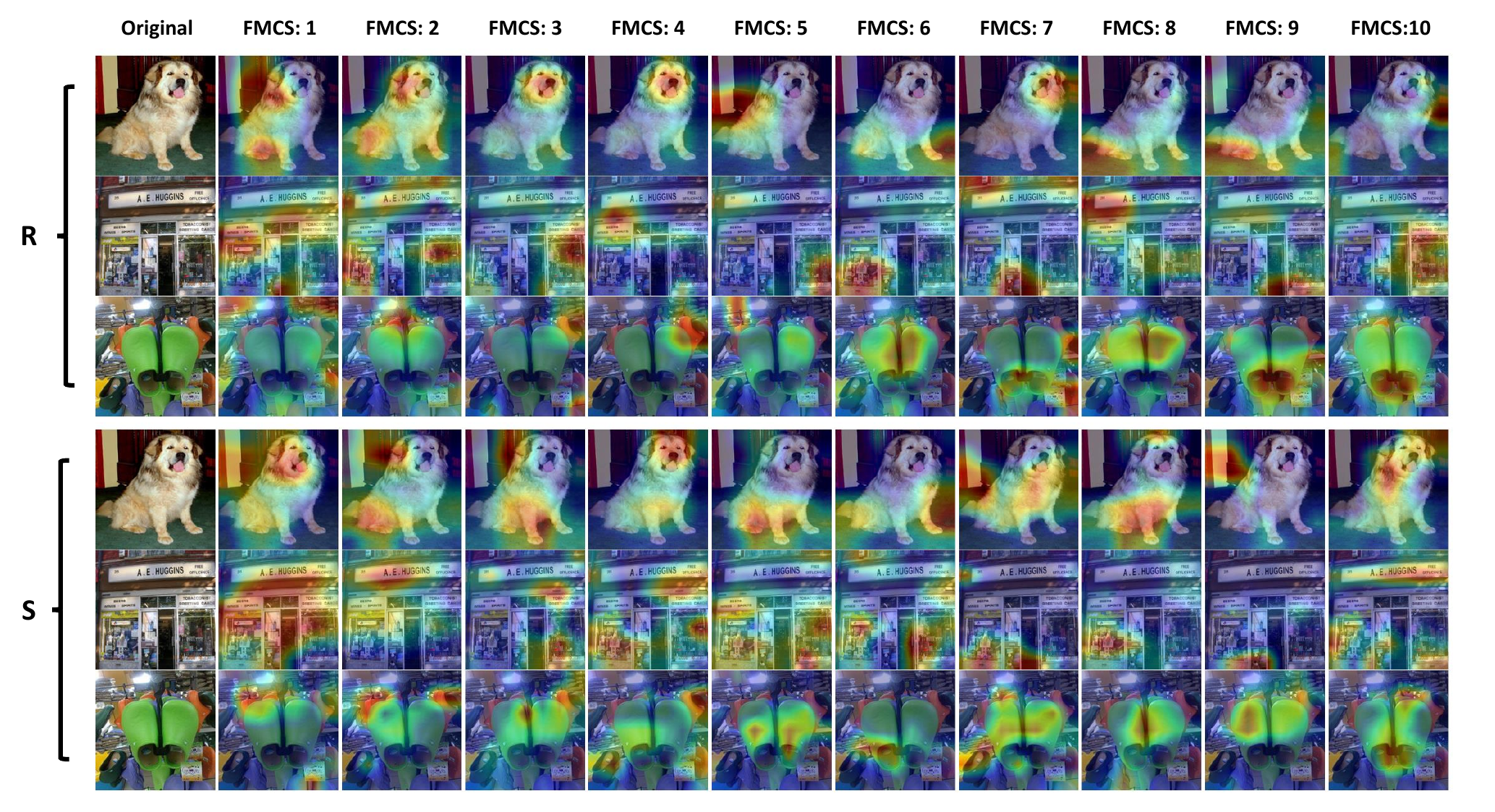}
    \captionsetup{font={small,stretch=1.1},justification=raggedright}
  \caption{\textbf{Grad-CAM Heatmaps Across 10 Convergence phases of Mini-Imagenet. }  The FMCS predictions under this dataset reach an average high level, demonstrating the efficiency of our methods on more complex classification tasks. }
  \label{fig:mini-image}
  \vspace{-0.3cm}
\end{figure*}

\subsection{Experiment Results}
 In this subsection, we will assess the effectiveness and practicality of our evaluation method from two perspectives: the effect of FMCS on the evaluation of functional module maturity and the predictive metrics of FMCE-Net for FMCS. 
 
To assess the effectiveness of FMCS, we employ the Grad-CAM method for visualizing feature maps across 10 convergence phases. Grad-CAM operates by analyzing the gradients of the last convolutional layer to identify the most crucial regions within the input image for predicting a specific class. 

In the heatmap of Class Activation Mapping (CAM), the shade of the color is generally proportional to the extent of the area's influence on the model's predictive outcome. And the color red typically signifies a positive contribution from that area to the model, meanwhile, the color blue represents that the features in those areas are unfavorable for the model's prediction of the current category. 
As can be observed from Fig.\ref{fig:mnist}, Fig.\ref{fig:cifar}, Fig.\ref{fig:mini-image}, the focus of the feature maps gradually shifts as the convergence degree increasing. In areas where the FMCS is low, the red regions tend to be larger and less pronounced, often concentrating along the edges of the target object. This suggests that the focus of the backbone has yet to hone in on the critical pixels. As the FMCS increases, the red regions become more concentrated on the key features of the target object, and the coloration becomes more vivid. This indicates an enhancement in the model's feature extraction capabilities. Visualizations confirm that FMCS, the quantitative indicator of convergence degree we proposed, can effectively evaluate the maturity of functional modules.

The predictive outcomes of the FMCE-Net on the FMCS-Datasets generated from six distinct dataset-backbone combinations are delineated in TABLE \ref{tab:FMCS prediction}.  An exhilarating finding is that the FMCE-Net has secured impressive performance metrics, with the average accuracy, precision, recall, and F1 score all exceeding \textit{97.5\%}  across the board. Notably, on the MNIST and Mini-ImageNet datasets, the predictive metrics of FMCE-Net are soared beyond the \textit{99\%} mark.
Additionally, these metrics achieve a high predictive level with low variances around \textit{3.5\%}, which demonstrates the effectiveness and robustness our evaluation method. The consistent high performance across different FMCS-datasets confirms excellent generalization capabilities in assessing feature map convergence within deep neural networks. 

We note that the FMCS prediction accuracy on the CIFAR-10 dataset is significantly lower compared to the other two datasets. By analyzing the heatmaps generated by Grad-CAM, it seems that backbones' focus are not directed towards the critical features in some images. Specifically, in the first three rows of Fig \ref{fig:cifar}, ResNet-50's attention is not concentrated on the key features. And depicted in the sixth row of Figure \ref{fig:cifar}, which presents an image annotated with 'dog',  ShuffleNet v2's focus is drawn to the background rather than the subject. This indicates that neither of the two backbone networks effectively leveraged their feature extraction capabilities on this dataset, leading to a confusion in the representation of feature maps across different convergence phases and consequently, a decrease in the predictive accuracy of FMCE-Net.

\section{CONCLUSIONS}
Existing evaluation methods for perceptual DNN models are end-to-end, lacking independent assessment of internal functional modules. This deficiency makes it challenging to interpret the models and optimize the functional modules accordingly. In this work, we are dedicated to conduct an  evaluation for the training maturity of functional modules, proposing an evaluation method for assessing the convergence of output feature maps. We introduce Convergence Quantitative Indicator (CQI) and Convergence threshold $\mu_{CQI}$, providing a more objective and precise assessment of model convergence. Then we develop a  evaluation metric Feature Map Convergence Score (FMCS), served as a standardized measure for quantifying the convergence phase of feature maps. Multiple FMCS-Datasets are generated to support the training of FMCE-Net, which designed to evaluate feature extraction module (backbone) in image classification tasks. Our evaluation method is verified to be effective and robust in assessing the convergence of feature maps during experiments, providing the deep learning field with a novel quantitative tool to measure and analyze the training efficiency and effectiveness of key components within DNN models. Future work will involve developing independent evaluation schemes for serially connected functional modules in more complex models, such as BEVFormer.

\addtolength{\textheight}{-12cm}   






\bibliographystyle{IEEEtran}
\bibliography{IEEEabrv,reference}

\begin{thebibliography}{10}
\providecommand{\url}[1]{#1}
\csname url@rmstyle\endcsname
\providecommand{\newblock}{\relax}
\providecommand{\bibinfo}[2]{#2}
\providecommand\BIBentrySTDinterwordspacing{\spaceskip=0pt\relax}
\providecommand\BIBentryALTinterwordstretchfactor{4}
\providecommand\BIBentryALTinterwordspacing{\spaceskip=\fontdimen2\font plus
\BIBentryALTinterwordstretchfactor\fontdimen3\font minus \fontdimen4\font\relax}
\providecommand\BIBforeignlanguage[2]{{%
\expandafter\ifx\csname l@#1\endcsname\relax
\typeout{** WARNING: IEEEtran.bst: No hyphenation pattern has been}%
\typeout{** loaded for the language `#1'. Using the pattern for}%
\typeout{** the default language instead.}%
\else
\language=\csname l@#1\endcsname
\fi
#2}}

\bibitem{kuznietsov2024explainable}
A.~Kuznietsov, B.~Gyevnar, C.~Wang, S.~Peters, and S.~V. Albrecht, ``Explainable ai for safe and trustworthy autonomous driving: A systematic review,'' \emph{arXiv preprint arXiv:2402.10086}, 2024.

\bibitem{zablocki2022explainability}
{\'E}.~Zablocki, H.~Ben-Younes, P.~P{\'e}rez, and M.~Cord, ``Explainability of deep vision-based autonomous driving systems: Review and challenges,'' \emph{International Journal of Computer Vision}, vol. 130, no.~10, pp. 2425--2452, 2022.

\bibitem{yang2023bevformer}
C.~Yang, Y.~Chen, H.~Tian, C.~Tao, X.~Zhu, Z.~Zhang, G.~Huang, H.~Li, Y.~Qiao, L.~Lu, \emph{et~al.}, ``Bevformer v2: Adapting modern image backbones to bird's-eye-view recognition via perspective supervision,'' in \emph{Proceedings of the IEEE/CVF Conference on Computer Vision and Pattern Recognition}, 2023, pp. 17\,830--17\,839.

\bibitem{pan2024clip}
C.~Pan, B.~Yaman, S.~Velipasalar, and L.~Ren, ``Clip-bevformer: Enhancing multi-view image-based bev detector with ground truth flow,'' \emph{arXiv preprint arXiv:2403.08919}, 2024.

\bibitem{andreas2016neural}
J.~Andreas, M.~Rohrbach, T.~Darrell, and D.~Klein, ``Neural module networks,'' in \emph{Proceedings of the IEEE conference on computer vision and pattern recognition}, 2016, pp. 39--48.

\bibitem{patrick2022capsule}
M.~K. Patrick, A.~F. Adekoya, A.~A. Mighty, and B.~Y. Edward, ``Capsule networks--a survey,'' \emph{Journal of King Saud University-computer and information sciences}, vol.~34, no.~1, pp. 1295--1310, 2022.

\bibitem{fernando2017pathnet}
C.~Fernando, D.~Banarse, C.~Blundell, Y.~Zwols, D.~Ha, A.~A. Rusu, A.~Pritzel, and D.~Wierstra, ``Pathnet: Evolution channels gradient descent in super neural networks,'' \emph{arXiv preprint arXiv:1701.08734}, 2017.

\bibitem{park2018accelerating}
K.~Park and D.-H. Kim, ``Accelerating image classification using feature map similarity in convolutional neural networks,'' \emph{Applied Sciences}, vol.~9, no.~1, p. 108, 2018.

\bibitem{azam2023using}
S.~Azam, S.~Montaha, K.~U. Fahim, A.~R.~H. Rafid, M.~S.~H. Mukta, and M.~Jonkman, ``Using feature maps to unpack the cnn ‘black box’theory with two medical datasets of different modality,'' \emph{Intelligent Systems with Applications}, vol.~18, p. 200233, 2023.

\bibitem{rahman2017feature}
I.~M. Rahman, C.~Hollitt, and M.~Zhang, ``Feature map quality score estimation through regression,'' \emph{IEEE Transactions on Image Processing}, vol.~27, no.~4, pp. 1793--1808, 2017.

\bibitem{chen2021feature}
W.~Chen, C.~Cong, and L.~Huang, ``Feature purity: A quantitative evaluation metric for feature extraction in convolutional neural networks,'' in \emph{Journal of Physics: Conference Series}, vol. 2010, no.~1.\hskip 1em plus 0.5em minus 0.4em\relax IOP Publishing, 2021, p. 012033.

\bibitem{zhao2021quantitative}
Y.~Zhao and H.~Zhang, ``Quantitative performance assessment of cnn units via topological entropy calculation,'' \emph{arXiv preprint arXiv:2103.09716}, 2021.

\bibitem{selvaraju2017grad}
R.~R. Selvaraju, M.~Cogswell, A.~Das, R.~Vedantam, D.~Parikh, and D.~Batra, ``Grad-cam: Visual explanations from deep networks via gradient-based localization,'' in \emph{Proceedings of the IEEE international conference on computer vision}, 2017, pp. 618--626.

\bibitem{simonyan2013deep}
K.~Simonyan, A.~Vedaldi, and A.~Zisserman, ``Deep inside convolutional networks: Visualising image classification models and saliency maps,'' \emph{arXiv preprint arXiv:1312.6034}, 2013.

\bibitem{chattopadhay2018grad}
A.~Chattopadhay, A.~Sarkar, P.~Howlader, and V.~N. Balasubramanian, ``Grad-cam++: Generalized gradient-based visual explanations for deep convolutional networks,'' in \emph{2018 IEEE winter conference on applications of computer vision (WACV)}.\hskip 1em plus 0.5em minus 0.4em\relax IEEE, 2018, pp. 839--847.

\bibitem{mankodiya2022od}
H.~Mankodiya, D.~Jadav, R.~Gupta, S.~Tanwar, W.-C. Hong, and R.~Sharma, ``Od-xai: Explainable ai-based semantic object detection for autonomous vehicles,'' \emph{Applied Sciences}, vol.~12, no.~11, p. 5310, 2022.

\bibitem{zheng2022shap}
Q.~Zheng, Z.~Wang, J.~Zhou, and J.~Lu, ``Shap-cam: Visual explanations for convolutional neural networks based on shapley value,'' in \emph{European Conference on Computer Vision}.\hskip 1em plus 0.5em minus 0.4em\relax Springer, 2022, pp. 459--474.

\bibitem{shi2020self}
W.~Shi, G.~Huang, S.~Song, Z.~Wang, T.~Lin, and C.~Wu, ``Self-supervised discovering of interpretable features for reinforcement learning,'' \emph{IEEE Transactions on Pattern Analysis and Machine Intelligence}, vol.~44, no.~5, pp. 2712--2724, 2020.

\bibitem{wang2021shapley}
R.~Wang, X.~Wang, and D.~I. Inouye, ``Shapley explanation networks,'' \emph{arXiv preprint arXiv:2104.02297}, 2021.

\bibitem{wang2019end}
D.~Wang, J.~Wen, Y.~Wang, X.~Huang, and F.~Pei, ``End-to-end self-driving using deep neural networks with multi-auxiliary tasks,'' \emph{Automotive Innovation}, vol.~2, pp. 127--136, 2019.

\bibitem{kim2022boosting}
Y.~Kim, S.~Kim, S.~Sim, J.~W. Choi, and D.~Kum, ``Boosting monocular 3d object detection with object-centric auxiliary depth supervision,'' \emph{IEEE Transactions on Intelligent Transportation Systems}, vol.~24, no.~2, pp. 1801--1813, 2022.

\bibitem{zheng2023gconet+}
P.~Zheng, H.~Fu, D.-P. Fan, Q.~Fan, J.~Qin, Y.-W. Tai, C.-K. Tang, and L.~Van~Gool, ``Gconet+: A stronger group collaborative co-salient object detector,'' \emph{IEEE Transactions on Pattern Analysis and Machine Intelligence}, 2023.

\end{thebibliography}

\end{document}